\pdfoutput=1
\documentclass[11pt]{article}

\usepackage[]{EMNLP2022}

\usepackage{times}
\usepackage{latexsym}

\usepackage[T1]{fontenc}
\usepackage[utf8]{inputenc}

\usepackage{microtype}

\usepackage{inconsolata}

\usepackage{enumitem}
\usepackage[labelformat=simple]{subcaption}

\usepackage{graphicx}
\usepackage{amsmath}
\usepackage{verbatim}
\usepackage{algorithm}
\usepackage{algpseudocode}
\usepackage{amsmath}
\usepackage{amssymb}
\usepackage{multirow}
\usepackage{makecell}
\usepackage{booktabs}
\usepackage{stfloats}
\usepackage[misc]{ifsym}

\title{Aligning Recommendation and Conversation via Dual Imitation}

\author{Jinfeng Zhou\textsuperscript{\rm 1,2}, Bo Wang\textsuperscript{\rm 1,2,\Letter}, Minlie Huang\textsuperscript{\rm 3}, Dongming Zhao\textsuperscript{\rm 4}, Kun Huang\textsuperscript{\rm 4}, \\
    {\bf Ruifang He\textsuperscript{\rm 1,2}, Yuexian Hou\textsuperscript{\rm 1}} \\
    \small \textsuperscript{\rm 1}College of Intelligence and Computing, Tianjin University, Tianjin, China \\
    \small \textsuperscript{\rm 2}State Key Laboratory of Communication Content Cognition, People's Daily Online, Beijing, China \\
    \small \textsuperscript{\rm 3}The CoAI Group, DCST, Institute for Artificial Intelligence, State Key Lab of Intelligent Technology and Systems, \\
    \small \textsuperscript{\rm 3}Beijing National Research Center for Information Science and Technology, Tsinghua University, Beijing 100084, China \\
    \small \textsuperscript{\rm 4}AI Lab, China Mobile Communication Group Tianjin Co., Ltd.\\
    \small \texttt{\{jfzhou, bo\_wang\}@tju.edu.cn \quad aihuang@tsinghua.edu.cn}}

\begin{document}
\maketitle

\begin{abstract}

Human conversations of recommendation naturally involve the shift of interests which can align the recommendation actions and conversation process to make accurate recommendations with rich explanations. However, existing conversational recommendation systems (CRS) ignore the advantage of user interest shift in connecting recommendation and conversation, which leads to an ineffective loose coupling structure of CRS. To address this issue, by modeling the recommendation actions as recommendation paths in a knowledge graph (KG), we propose DICR (\textbf{D}ual \textbf{I}mitation for \textbf{C}onversational \textbf{R}ecommendation), which designs a dual imitation to explicitly align the recommendation paths and user interest shift paths in a recommendation module and a conversation module, respectively. By exchanging alignment signals, DICR achieves bidirectional promotion between recommendation and conversation modules and generates high-quality responses with accurate recommendations and coherent explanations. Experiments demonstrate that DICR outperforms the state-of-the-art models on recommendation and conversation performance with automatic, human, and novel explainability metrics.
\end{abstract}

\section{Introduction}

\begin{figure}[t]
\centering
\includegraphics[width=0.90\columnwidth]{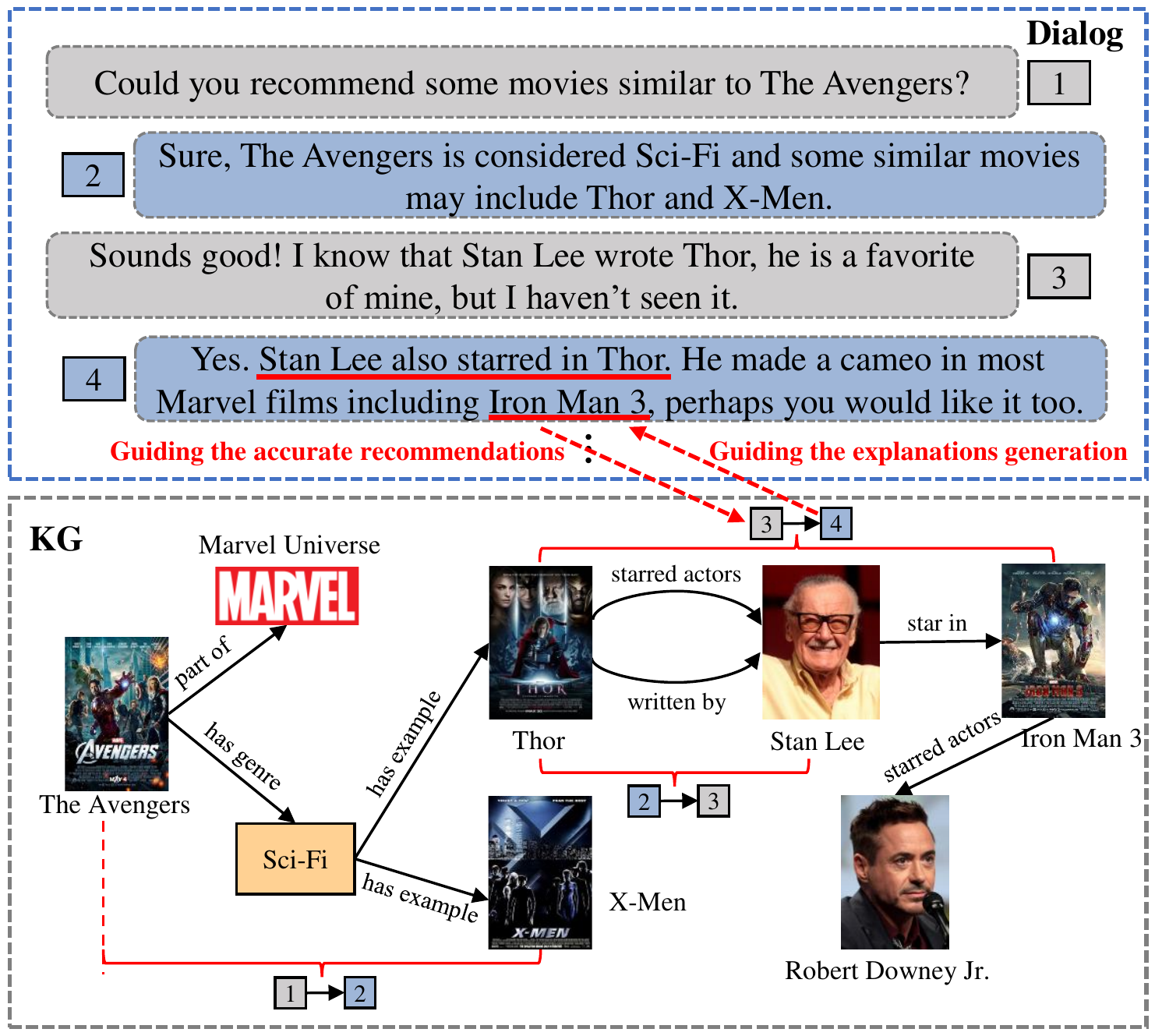} % Reduce the figure size so that it is slightly narrower than the column. Don't use precise values for figure width.This setup will avoid overfull boxes.
\caption{The interest shift process expressed in the conversation can guide the generation of explainable recommendation path. The explainable recommendation path, in turn, can guide the generation of explainable response containing accurate recommendations. Recommendation and conversation maximize mutual benefits in bidirectional guidance.}
\label{introduction}
\end{figure}

Conversational recommendation systems (CRS) \cite{DBLP:conf/acl/LiuWNWCL20,li2022user} aim to conduct recommendations during conversations with users \cite{GAO2021100}. Compared with traditional recommendation systems \cite{DBLP:conf/aaai/WangWX00C19,DBLP:conf/sigir/XianFMMZ19}, CRS’s two main advantages are understanding user’s dynamic interest during the conversation and making persuasive response with coherent explanations of the recommendation \cite{DBLP:journals/csur/JannachMCC21}. In both advantages, user interest shift plays an essential role. As the dialog in Fig. \ref{introduction}, the successful recommendation of “\textsl{Iron Man 3}” in the final response is achieved by tracking and reasoning the user interest shift of “\textsl{The Avengers$\rightarrow$Sci-Fi$\rightarrow$Thor$\rightarrow$Stan Lee$\rightarrow$Iron Man 3}”. Furthermore, the final response is persuasive because of utilizing part of the user interest shift, i.e., “\textsl{Thor$\rightarrow$Stan Lee$\rightarrow$Iron Man 3}”, as the explanation.

Due to the limited context \cite{DBLP:conf/emnlp/HayatiKZSY20}, a recommendation module based on knowledge graph (KG)
is helpful to track the user interest shift in conversation. As shown in Figure \ref{introduction}, formally corresponding with the paths in KG, the user interest shift in conversation not only guides the reasoning-based prediction of recommendation, but also guides the explanation generation in response.

However, existing KG-enhanced CRS models \cite{DBLP:conf/emnlp/ChenLZDCYT19,DBLP:conf/kdd/ZhouZBZWY20,DBLP:conf/emnlp/ZhouWHH21,DBLP:journals/corr/abs-2201-02732,DBLP:conf/acl-convai/ZhangLLZZWM22} have not made full use of the user interest shift to tightly align the KG-based recommendation and conversation. Consequently, one issue is the less accurate recommendation due to using unrelated entities in conversation to support the recommendation instead of using coherent entities in user interest shift. The other issue is the lack of explanation in response due to black-box representation of the user preference ignoring the explicit preference logic in user interest shift.

To address these issues in CRS, we propose to align the explicit behaviors of recommendation reasoning and conversation process, which are described as the recommendation paths and interest shift paths, respectively. As in Figure \ref{introduction}, a \textbf{recommendation path} is an explicit path in KG consisting of explicit relations of entity nodes and ending with a predicted recommended entity; an \textbf{interest shift path} is an implicit path in dialog context consisting of implicit relations of entity words. The recommendation path and interest shift path are concrete manifestations of the user interest shift in KG and dialog, respectively. The sequence of interest entities shared by the two paths facilitates the alignment of recommendation reasoning and conversation process, which can be effectively achieved by imitation learning \cite{DBLP:conf/nips/HoE16}. Therefore, we propose a dual imitation framework named DICR (\textbf{D}ual \textbf{I}mitation for \textbf{C}onversational \textbf{R}ecommendation). DICR designs bidirectional alignment signals from dual imitation learning to improve the CRS by forcing the recommendation and conversation to behave similarly to the shared user interest shift. 

Precisely, in a conversation-aware recommendation module, to align the recommendation reasoning to the conversational user interest shift, the recommendation side of the dual imitation, i.e., path imitation, adopts adversarial reinforcement learning to make the recommendation reasoning policy imitate the user interest shift in conversation. The reasoned recommendation paths are provided to the conversation module as alignment signals. In a recommendation-aware conversation module, to align the conversation process to the recommendation paths, the conversation side of the dual imitation, i.e., knowledge imitation and semantic imitation, which refine weight distribution and semantic encoding of recommendation paths by imitating the human response and the utterance statement semantic of golden explanations, respectively. These two imitations also provide the recommendation module with the rewards as the alignment signals indicating how well the predicated recommendation paths consist to the conversation context.

Our contributions are summarized as follows:

(1) To the best of our knowledge, we are the first to adopt imitation learning in CRS to integrate recommendation and conversation tightly. We design a dual imitation framework named DICR, which aligns recommendation and conversation behavior and promotes bidirectional improvement, taking recommendation paths and conversational rewards from the dual imitation as alignment signals.

(2) The dual imitation benefits the knowledge acquisition and semantic generation, promoting the accuracy of recommendation and significantly improving the explanations of recommendations in generated responses with coherent knowledge. 
 
(3) Extensive experiments demonstrate that DICR outperforms the SOTA models on both recommendation and conversation performance with automatic, human and novel explainability metrics. 

\section{Related Work}

Conversational recommendation systems (CRS) aim to obtain user interests through conversational interaction and make persuasive recommendations \cite{DBLP:journals/csur/JannachMCC21, GAO2021100, DBLP:conf/sigir/RenYCWH021}. To track the user interest shift in conversation, an intuitive strategy is to ask related questions \cite{DBLP:conf/recsys/KostricBR21, DBLP:conf/www/ZhangWSPWXLP22} which leads to question-based CRS \cite{DBLP:conf/kdd/LeiZ0MWCC20, DBLP:conf/sigir/DengL0DL21}. Limited by predefined templates for asking and recommending, it is difficult for question-based CRS to flexibly adopt different contexts and converse in a human-like manner.

Towards more flexible conversation, generation-based CRS \cite{DBLP:conf/nips/LiKSMCP18, DBLP:conf/emnlp/ZhangYC0Y21, DBLP:conf/emnlp/LiangHXMHCGLJ21} capture user interests from context and generate responses containing persuasive explanations for recommendation. Limited by sparse context and language complexity \cite{DBLP:conf/acl/LuBSMCWH21, yang-etal-2022-topkg}, it is challenging for generation-based CRS to track user interest shift in conversation. As a popular solution, KG-enhanced CRS \cite{DBLP:conf/acl/MoonSKS19, DBLP:conf/emnlp/MaTH21, zhou-etal-2022-cr} involve knowledge graphs about explicit relations among potential interest items. 

Although KG-enhanced CRS has achieved significant improvements, most approaches \cite{DBLP:conf/emnlp/ChenLZDCYT19,DBLP:conf/kdd/ZhouZBZWY20,DBLP:conf/emnlp/ZhouWHH21,DBLP:journals/corr/abs-2201-02732} adopt a black-box style to transfer implicit and sparse information between recommendation and conversation. The explicit interest reasoning path in KG and explicit interest shift path in conversation have a good chance to align the recommendation and conversation to benefit each other.

\begin{figure*}[t]
\centering
\includegraphics[width=0.9\textwidth]{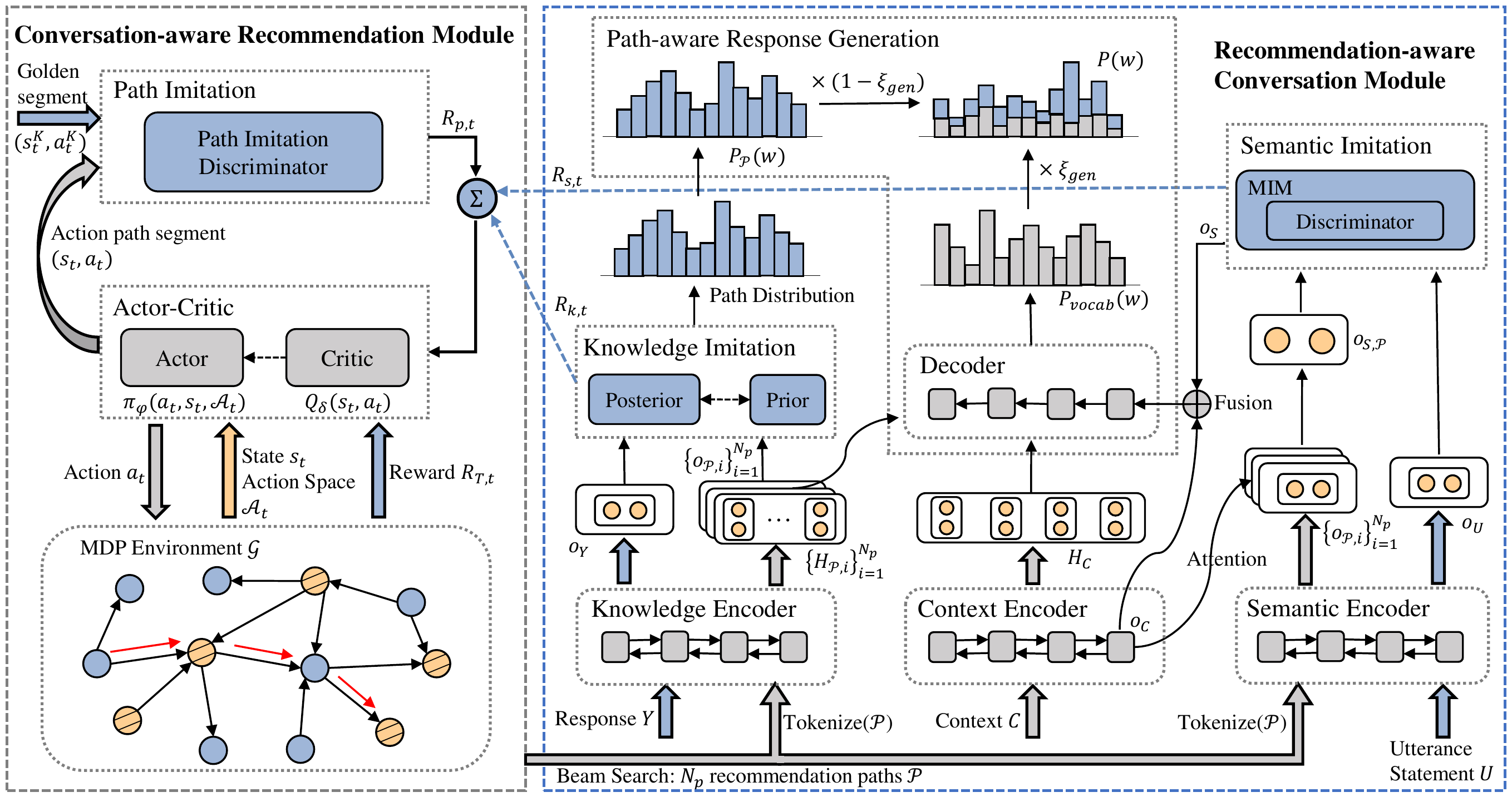} % Reduce the figure size so that it is slightly narrower than the column.
\caption{The dual imitation architecture of the proposed DICR model.}
\label{framework}
\end{figure*}

\section{Problem Formalization}

A KG $\mathcal{G}=\left\{\left(e, r, e^{\prime}\right) \mid e, e^{\prime} \in \mathcal{E}, r \in \mathcal{R}\right\}$, where $\mathcal{E}$ is entity set, and $\mathcal{R}$ is relation set. Each triplet $(e,r,e^{\prime})$ indicates that the head entity $e$ and the tail entity $e^{\prime}$ are connected by the relation $r$. In this paper, a recommendation path is a multi-hop reasoning path $p$ on $\mathcal{G}$: $p=\left\{e_{0} \stackrel{r_{1}}{\rightarrow} e_{1} \stackrel{r_{2}}{\rightarrow} \ldots \stackrel{r_{t}}{\rightarrow} e_{t}\right\}$.

Suppose we have a conversational recommendation corpus $\mathcal{D}$ parallel to a knowledge graph $\mathcal{G}$, in which the interests (e.g., movies) mentioned in $\mathcal{D}$ are linked to the entities in $\mathcal{G}$. $C=\left(c_{1}, c_{2}, \ldots, c_{n}\right)$ is the conversation context, where $c_{i}$ is an utterance. $I$ is the set of recommendation items under $C$. $Y=\left(y_{1}, y_{2}, \ldots, y_{m}\right)$ is a response containing $I$, where $y_{i}$ is a token. $K=\left\{e_{0} \stackrel{r_{1}^{K}}{\rightarrow} e_{1} \stackrel{r_{2}^{K}}{\rightarrow} \ldots \stackrel{r_{l}^{K}}{\rightarrow} e_{l}\right\}$ is a golden interest shift path connecting the interest entities $e_{0,1,...,l}$ in $C$ and $Y$. $K$ also matches a recommendation path $p$ in $\mathcal{G}$. $K$ can be extracted by identifying entities in the conversation and linking the entities to the nodes in $\mathcal{G}$. The logical utterance statement of $K$ is $U$, which is the explanation of recommendation, e.g., given a one-hop reasoning path in $\mathcal{G}$ (“\textit{Thor}”, “\textit{written\_by}”, “\textit{Stan Lee}”), its tokenized $U$ can be “\textit{Thor is written by Stan Lee.}”.

In this paper, given a conversation context $C$ and a KG $\mathcal{G}$, we aim to generate a response $Y$ containing the recommendation set $I$ and explanation $U$. We design a novel CRS model, in which KG path reasoning obtains explicit reasoning path set $\mathcal{P}$ from $\mathcal{G}$ to help generate $Y$ containing recommendation set $I$ and coherent explanation $U$ of $I$.

\section{Approach}

\subsection{Architecture Overview}

As shown in Fig. \ref{framework}, in DICR, the \textbf{conversation-aware recommendation module} learns recommendation path reasoning policy with adversarial reinforcement learning. The path imitation discriminator aligns the recommendation paths with the golden interest shift path to reward the agent $R_{p,t}$ to optimize the reasoning policy. As a result, the top tokenized recommendation paths are provided to the conversation module as alignment signals. In the \textbf{recommendation-aware conversation module}, the knowledge imitation aligns the prior and posterior recommendation knowledge in tokenized recommendation paths and human response, respectively. The semantic imitation uses  Mutual Information Maximization (MIM) to align the semantic encoding of recommendation paths with that of the utterance statement of the golden interest shift path. These imitations refine distribution of knowledge and overall words and thus benefit the path-aware response generation. They also generate rewards $R_{k,t}$ and $R_{s,t}$ as alignment signals to guide the recommendation path reasoning. Finally, DICR performs a joint training to bidirectionally promote the recommendation and conversation with alignment signals from dual imitation. 

\subsection{Conversation-aware Recommendation Module}

In this module, we formalize the user interest shift in conversation as a Markov Decision Process (MDP) \cite{sutton2018reinforcement} through KG paths to reason interest shift path with adversarial reinforcement learning. We construct KG embeddings \cite{DBLP:conf/nips/BordesUGWY13} of each entity. The entities mentioned in $C$ are extracted by fuzzy matching, whose embeddings are averaged as the preference representation of user $u$ in the current context. 

\textbf{State}. We start path reasoning from the starting entity $e_{0}$ in $C$.  The initial state $s_{0} \in \mathcal{S}$ is $s_{0}=\left\{u,e_{0}\right\}$. We encode the $H$-step history of entities and relations as the observed state $s_{t} \in \mathcal{S}$ at step t, i.e., $s_{t}=\left\{u, e_{t-H}, \ldots, r_{t}, e_{t}\right\}$, whose embedding $\boldsymbol{s}_{t}$ is obtained by concatenating the 
embeddings of all members of $s_{t}$, i.e., $\boldsymbol{s}_{t}=\boldsymbol{u} \oplus \boldsymbol{e}_{t-H} \ldots \oplus \boldsymbol{r}_{t} \oplus \boldsymbol{e}_{t}$, where $\boldsymbol{u}$ is preference representation, $\oplus$ is the concatenation operator. If the path length is smaller than $H$, we pad $\boldsymbol{s}_{t}$ with zeros.

\textbf{Action}. The action space $\mathcal{A}_{t}$ of the state $s_{t}$ is defined as all outgoing edges of the entity $e_{t}$ in the KG $\mathcal{G}$, excluding history entities and relations. $\mathcal{A}_{t}=\left\{(r, e) \mid\left(e_{t}, r, e\right) \in \mathcal{G}, e \notin\left\{e_{0}, \ldots, e_{t-1}\right\}\right\}$. As an option to terminate, $\mathcal{A}_{t}$ has a self-loop edge.

\textbf{Transition}. Given the current state $s_{t}$ and the action chosen by the agent $a_{t}=\left(r_{t+1}, e_{t+1}\right)$, the next state $s_{t+1}$ is: $s_{t+1}=\mathcal{T}\left(s_{t}, a_{t}\right)=\left\{u, e_{t-H+1}, \ldots, r_{t+1}, e_{t+1}\right\}$, where $\mathcal{T}: \mathcal{S} \times \mathcal{A} \rightarrow \mathcal{S}$ refers to the state transition function.

\textbf{Reward}. We only give the agent terminal reward $R_{T,t}$, $R_{T,t}$ is 1 if the agent generates a path ends with the recommended items $I_{Y}$ in the response $Y$, and 0 otherwise. 

\textbf{Policy Optimization}. We adopt adversarial imitation learning \cite{DBLP:conf/sigir/ZhaoWZZLX020} based on the Actor-Critic method for policy optimization. Actor learns a path reasoning policy $\pi_{\varphi}\left(a_{t}, s_{t}, \mathcal{A}_{t}\right)$ which selects a “good” action $a_{t}$ based on the current state $s_{t}$ and the action space $\mathcal{A}_{t}$ to “fool” the discriminator in the path imitation. Critic estimates the value $Q_{\delta}\left(s_{t}, a_{t}\right)$ of each action $a_{t}$ in the situation of the state $s_{t}$ to guide the actor to choose a “good” action. We use two fully connected layers as the actor policy network:
\begin{equation}
\pi_{\varphi}\left(a_{t}, s_{t}, \mathcal{A}_{t}\right)=\eta\left(\boldsymbol{A}_{t} f\left(W_{\varphi, 2}\left(f\left(W_{\varphi, 1} \boldsymbol{s}_{t}\right)\right)\right)\right),
\end{equation}
where $\boldsymbol{A}_{t}$ denotes that the action space is encoded by stacking the embedding of all actions in $\mathcal{A}_{t}$, and each embedding in $a_{t} \in \mathcal{A}_{t}$ is obtained by a lookup layer. $\eta(\cdot)$ is the softmax function, $f(\cdot)$ is the ELU activation function, $W_{\varphi, 1}$ and $W_{\varphi, 2}$ are learnable. We design a critic network as:
%similar to the actor policy network:
\begin{equation}
Q_{\delta}\left(s_{t}, a_{t}\right)=\boldsymbol{a}_{\delta, t} \cdot f\left(W_{\delta, 2}\left(f\left(W_{\delta, 1} \boldsymbol{s}_{t}\right)\right)\right),
\end{equation}
where $\boldsymbol{a}_{\delta, t}$ is the embedding of action $a_{t}$ in the critic, $f(\cdot)$ is the ELU activation function, $W_{\delta, 1}$ and $W_{\delta, 2}$ are learnable.

\paragraph{Path Imitation} To guide the actor to generate a path in line with the user interest shift, we design the path imitation discriminator $\mathcal{I}_{p, \tau}$, which judges whether the path segment generated by the actor at each step t is similar to the golden interest shift path segment in current context. Given the current state $s_{t}$ and action $a_{t}$, the probability that $\mathcal{I}_{p, \tau}$ outputs $\left(s_{t}, a_{t}\right)$ conforms to the golden shift path segment:
\begin{equation}
\mathcal{I}_{p, \tau}\left(s_{t}, a_{t}\right)=\sigma(\boldsymbol{b}_{p, \tau}^{T} z\left(W_{p, \tau} z\left(\boldsymbol{s}_{t} \oplus \boldsymbol{a}_{p, t}\right)\right)),
\end{equation}
where $\boldsymbol{a}_{p, t}$ is the embedding of $a_{t}$ in $\mathcal{I}_{p, \tau}$, $z(\cdot)$ is tanh function and $\sigma(\cdot)$ is sigmoid function. $W_{p, \tau}$ and $\boldsymbol{b}_{p, \tau}$ are learnable. $\mathcal{I}_{p, \tau}$ is learned by minimizing the loss function $L_{\tau}$:
% \begin{small}
\begin{equation}
L_{\tau}=-(\log \left(1-\mathcal{I}_{p, \tau}\left(s_{t}, a_{t}\right)\right)+\log (\mathcal{I}_{p, \tau}(s_{t}^{K}, a_{t}^{K})))
\end{equation}
% \end{small}
where $s_{t}^{K}$ and $a_{t}^{K}$ respectively denote the state and action of the golden shift process in the same step t. We further obtain the reward $R_{p, t}$ given by $\mathcal{I}_{p, \tau}$ to the actor at each step t:
\begin{equation}
R_{p, t}=\log \left(\mathcal{I}_{p, \tau}\left(s_{t}, a_{t}\right)\right)-\log \left(1-\mathcal{I}_{p, \tau}\left(s_{t}, a_{t}\right)\right).
\end{equation}

Here, the aggregation reward obtained by the agent is $R_{t}=\alpha R_{p, t}+(1-\alpha) R_{T, t}$. where $\alpha \in[0,1]$. In the final joint training with the conversation module, the agent receives two other rewards $R_{k,t}$ and $R_{s,t}$ from the conversation module. Given $Q_{\delta}\left(s_{t}, a_{t}\right)$, the actor and critic is updated jointly by minimizing the loss function $L_{\varphi, \delta}$:
\begin{equation}
L_{\varphi, \delta}=-\mathrm{E}_{a \sim \pi_{\varphi}} Q_{\delta}\left(s_{t}, a\right)+\left(Q_{\delta}\left(s_{t}, a_{t}\right)-G_{t}\right)^{2},
\end{equation}
where the total cumulative reward $G_{t}=R_{t}+\mathrm{E}_{a \sim \pi_{\varphi}} Q_{\delta}\left(s_{t+1}, a\right)$ is calculated by Bellman equation \cite{bellman2013dynamic}. The actor, critic and path imitation discriminator is jointly optimized by minimizing $L_{R E C}=L_{\varphi, \delta}+L_{\tau}$.

\paragraph{Beam Search of Recommendation Paths} After agent pre-training, we adopt beam search to generate candidate recommendation paths. Sorted by the probability of leading to accurate recommendation, top $N_{p}$ paths are tokenized into a statement containing entity and relation words, which are provided to the conversation module as alignment signals.

\subsection{Recommendation-aware Conversation Module}

\paragraph{Encoder}
The conversation context $C$, the response $Y$, the utterance $U$ and the tokenized recommendation paths $\mathcal{P}$ are encoded by the context encoder, the knowledge encoder and the semantic encoder, respectively, based on Bi-RNN. Given the input sequence $X=\left(x_{1}, \ldots, x_{N}\right)$, the forward and the backward RNN respectively generate hidden states $h_{t}^{f}$ and $h_{t}^{b}$ for each $x_{t}$, which are concatenated to form the overall hidden state $h_{t}$:
% \begin{small}
\begin{equation}
h_{t}=[h_{t}^{f} ; h_{t}^{b}]=[\overrightarrow{\operatorname{GRU}}(x_{t}, h_{t-1}^{f}) ; \overleftarrow{\operatorname{GRU}}(x_{t}, h_{t+1}^{b})],
\end{equation}
% \end{small}
where $[;]$ is the concatenation operation. We denote the hidden states of all time steps as $H=\left(h_{1}, h_{2}, \ldots, h_{N}\right)$, $o=\left[h_{N}^{f} ; h_{1}^{b}\right]$ as the final hidden state. For all input sources, we obtain $H_{C}$, $H_{Y}$, $H_{U}$, $\left\{H_{\mathcal{P}, i}\right\}_{i=1}^{N_{p}}$ and $o_{C}$, $o_{Y}$, $o_{U}$, $\left\{o_{\mathcal{P}, i}\right\}_{i=1}^{N_{p}}$.

\paragraph{Knowledge Imitation}

To refine the distribution of tokenized recommendation paths of leading to accurate recommendation with proper explanation, knowledge imitation makes the tokenized recommendation paths imitate the human response, which often contains correct recommendation destination without strong explanation. Given the encoded conversation context $o_{C}$ and $o_{\mathcal{P}}=\left\{o_{\mathcal{P}, i}\right\}_{i=1}^{N_{p}}$ of the encoding of tokenized recommendation paths, the $o_{C}$ server as the prior information. We first obtain the prior path weight distribution by the similarity between $o_{C}$ and each path $p_{i}$:
\begin{equation}
P\left(p_{i} \mid C\right)=\frac{\exp \left(o_{\mathcal{P}, i} \cdot z\left(W_{K, C} o_{C}\right)\right)}{\sum_{j=1}^{N_{p}} \exp \left(o_{\mathcal{P}, j} \cdot z\left(W_{K, C} o_{C}\right)\right)},
\end{equation}
where $z(\cdot)$ is the tanh function, $W_{K,C}$ is learnable. 

Since the recommendation paths contain the predicated interest in response, the prior information is insufficient to calculate the recommendation paths distribution. Therefore, the imitation also involves the human response as posterior information to obtain the posterior distribution of the paths.
\begin{equation}
P\left(p_{i} \mid Y\right)=\frac{\exp \left(o_{\mathcal{P}, i} \cdot z\left(W_{K, Y} o_{Y}\right)\right)}{\sum_{j=1}^{N_{p}} \exp \left(o_{\mathcal{P}, j} \cdot z\left(W_{K, Y} o_{Y}\right)\right)},
\end{equation}
where $W_{K,Y}$ is learnable. We use Kullback-Leibler divergence loss $L_{K L}$ to make $P\left(p_{i} \mid C\right)$ imitates  $P\left(p_{i} \mid Y\right)$, and BOW Loss $L_{B O W}$ to enforce the relevancy between recommendation paths distribution and response \cite{DBLP:conf/ijcai/LianXWPW19}. 

\paragraph{Semantic Imitation}
To refine the semantic encoding of tokenized recommendation paths, semantic imitation makes tokenized recommendation paths imitates the golden utterance statement of correct recommendation and coherent explanation. Given the encoded conversation context $o_{C}$ and the hidden state $o_{\mathcal{P}}=\left\{o_{\mathcal{P}, i}\right\}_{i=1}^{N_{p}}$ of the tokenized recommendation paths encoding, we apply attention \cite{DBLP:journals/corr/BahdanauCB14} to the $o_{\mathcal{P}}$ to obtain the context-based path aggregation representation $o_{S, \mathcal{P}}=\operatorname{Attention}\left(o_{\mathcal{P}}, z\left(W_{S, \mathcal{P}} o_{C}\right)\right)$,
%\begin{equation}
%o_{S, \mathcal{P}}=\operatorname{Attention}\left(o_{\mathcal{P}}, z\left(W_{S, %\mathcal{P}} o_{C}\right)\right),
%\end{equation}
where $z(\cdot)$ is the tanh function, $W_{S,\mathcal{P}}$ is the parameter matrix. 

To make $o_{S,\mathcal{P}}$ and the semantic $o_{U}$ of the encoded golden interest shift path behave similarly, we adopt the Mutual Information Maximization \cite{DBLP:conf/wsdm/CaoLGLL021}, which forces the learned context-based aggregation representation to equip with the semantic of the golden utterance statement via maximizing the mutual information between $o_{S,\mathcal{P}}$ and $o_{U}$. We use binary cross-entropy loss as the mutual information estimator. The learning objective is:
% \begin{equation}
\begin{align}
L_{B C E}&=-\frac{1}{|\mathbb{P}|+|\mathbb{N}|}\big(\sum\nolimits_{\mathbb{P}} \log \mathcal{I}_{S, \phi}\left(o_{S, \mathcal{P}}, o_{U}\right)+ \nonumber \\
&\sum\nolimits_{\mathbb{N}} \log \left(1-\mathcal{I}_{S, \phi}\left(\widetilde{o_{S, \mathcal{P}}}, o_{U}\right)\right)\big),
\end{align}
% \end{equation}
where $\mathbb{P}$ and $\mathbb{N}$ represent the set of positive and negative samples, respectively. $\widetilde{o_{S, \mathcal{P}}}$ is the random sampled negative sample's encoding. $\mathcal{I}_{S, \phi}$ is a semantic imitation discriminator to score $o_{S,\mathcal{P}}$ and $o_{U}$ via a bilinear mapping function:
\begin{equation}
\mathcal{I}_{S, \phi}\left(o_{S, \mathcal{P}}, o_{U}\right)=\sigma(\left(o_{S, \mathcal{P}}\right)^{T} W_{S, \phi} o_{U}),
\end{equation}
where $\sigma(\cdot)$ is the sigmoid function, $W_{S, \phi}$ is the parameter matrix. For the response generation, an MLP layer merges the learned semantic $o_{S}$ into the hidden state $o_{C}$ of the conversation context as the initial hidden state of the decoder, where $o_{S}=o_{U}$ if $U$ is available, otherwise $o_{S}$ is the learned $o_{S,\mathcal{P}}$. In the inference stage, $U$ is unknown. 

\paragraph{Path-aware Response Generation}
We employ GRU to integrate context and path information to generate a response. Given the decoder state $h_{t}$, the output states $H_{C}$ and $\left\{H_{\mathcal{P}, i}\right\}_{i=1}^{N_{p}}$ of the context encoder and knowledge encoder, we apply attention to the $H_{C}$ at decoder step t: $d_{t}^{C}, v_{t}^{C}=\operatorname{Attention}\left(H_{C}, h_{t}\right)$, where $d_{t}^{C}$ is the attention distribution of each token in the context $C$, $v_{t}^{C}$ is the aggregation vector of $C$. $\left\{d_{t}^{\mathcal{P}, i}\right\}_{i=1}^{N_{p}}$ and $\left\{v_{t}^{\mathcal{P}, i}\right\}_{i=1}^{N_{p}}$ are obtained for $H_{\mathcal{P},i}$ of each path $p_{i}$. We obtain the overall path representation $v_{t}^{\mathcal{P}}=\sum_{i=1}^{N_{p}} \mu_{\mathcal{P}, i} \cdot v_{t}^{\mathcal{P}, i}$, where $\mu_{\mathcal{P}, i}=P\left(p_{i} \mid Y\right)$ in training, 
and $\mu_{\mathcal{P}, i}=P\left(p_{i} \mid C\right)$ in inference. To reduce the impact of inaccurate recommended paths, we design a fusion gate $g_{t}$ to determine the contribution of $v_{t}^{\mathcal{P}}$ to the fusion information $v_{t}$:
\begin{equation}
\begin{gathered}
g_{t}=\sigma\left(W_{g}\left[d_{t}^{C} ; v_{t}^{\mathcal{P}}\right]\right), v_{t}=g_{t} d_{t}^{C}+\left(1-g_{t}\right) v_{t}^{\mathcal{P}},
\end{gathered}
\end{equation}
where $W_{g}$ is learnable. Hence. the decoder updates its state as: $h_{t+1}=\operatorname{GRU}\left(h_{t},\left[y_{t} ; v_{t}\right]\right)$, where $y_{t}$ is the embedding of predicted word at time step t. $h_{t}$ and $v_{t}$ are also used to obtain the generation probability $P_{vocab}\left(w_{t}\right)$ over the vocabulary at time step t, formalized as $P_{vocab }\left(w_{t}\right)=\rho\left(\left[h_{t} ; v_{t}\right]\right)$, where $\rho(\cdot)$ is a two-layer MLP with a softmax function. Furthermore, we adopt a pointer copy mechanism to copy tokens from the tokenized recommendation paths $\mathcal{P}$, which ensures that the logical knowledge in the path can be copied to enrich the explanation in the response. At time step t, the probability of copying tokens from $\mathcal{P}$ is a weighted sum of copying tokens from all paths over the path distribution:
\begin{align}
P_{\mathcal{P}}\left(w_{t}\right)=\sum_{i=1}^{N_{p}} \mu_{\mathcal{P}, i} \cdot \sum\nolimits_{\left\{j: p_{i}^{j}=w_{t}\right\}} d_{t, j}^{\mathcal{P}, i},
\end{align}
where $p_{i}^{j}$ is the token in the path $p_{i}$, $d_{t,j}^{\mathcal{P},i}$ is the attention weights of the $j^{th}$ token in $p_{i}$. We use a pointer generation probability $\xi_{t}^{gen}$ \cite{DBLP:conf/acl/SeeLM17} to obtain the overall probability distribution:
\begin{equation}
P\left(w_{t}\right)=\xi_{t}^{gen} P_{vocab}\left(w_{t}\right)+\left(1-\xi_{t}^{gen}\right) P_{\mathcal{P}}\left(w_{t}\right),
\end{equation}
where $\xi_{t}^{gen}=\sigma\left(W_{gen}\left[y_{t-1} ; h_{t} ; v_{t}\right]\right)$,  $W_{gen}$ is learnable. When training the conversation module, we use additional NLL Loss to quantify the difference between the golden and generated response:
\begin{equation}
L_{N L L}=-\frac{1}{|Y|} \sum_{t=1}^{|Y|} \log (P(y_{t} \mid y_{\textless t}, C, \mathcal{P})).
\end{equation}

In summary, the conversation module is jointly optimized by minimizing the joint loss $L_{G E N}=L_{K L}+L_{B O W}+L_{B C E}+L_{N L L}$.

\subsection{Bidirectional Improvement of Two Modules}
After training recommendation and conversation modules, we conduct a bidirectional joint training. The conversation module provides the recommendation module with the rewards $R_{k,t}$ and $R_{s,t}$ from the knowledge imitation and semantic imitation, respectively. $R_{k,t}$ is the knowledge consistency between recommendation paths and human response. If path $p \in \mathcal{P}$, $R_{k, t}=\log \left(\mu_{\mathcal{P}, i}\right)+\log \left(1-\mu_{\mathcal{P}, i}\right)$, where $i$ is the index of $p$ in $\mathcal{P}$, otherwise $R_{k,t}=0$. $R_{s,t}$ is the semantic similarity between the path segment generated at step $t$ and the golden utterance. $R_{s, t}=\log \left(\mathcal{I}_{S, \phi}\left(o_{p, t}, o_{U}\right)\right)+\log \left(1-\mathcal{I}_{S, \phi}\left(o_{p, t}, o_{U}\right)\right)$, where $o_{p,t}$ is the hidden state of the tokenized path segment encoded by the semantic encoder. The aggregation reward is $R_{t}=\alpha R_{p, t}+\beta R_{k, t}+\gamma R_{s, t}+(1-\alpha-\beta-\gamma) R_{T, t}$, where $\alpha+\beta+\gamma \in[0,1]$. If the path is 
shorter than the maximum reasoning length, $\beta=0$. 

The recommendation module provides the conversation module with optimized recommendation paths to guide the response generation. In this way, the bidirectional joint training optimizes the alignment between recommendation and conversation and promotes the overall performance of CRS.

\section{Experiments}

\subsection{Experiment Setup}

\paragraph{Dataset}
We did experiments on OpenDialKG \cite{DBLP:conf/acl/MoonSKS19}, a dialog$\leftrightarrow$KG parallel corpus for CRS, where the mentions of KG entities and their factual connections in a dialog are annotated. The user interest shift path is extracted from context-response pairs, where its start entity is in the context, and destination entity is in the response. Each path is tokenized into an utterance statement that weaves together the entities and relations mentioned in the conversation. More details on data and experiments are in Appendix \ref{sec:dataset} and \ref{sec:pathnumber}.

\paragraph{Models for Comparison}

(1) \textit{TextCNN} \cite{DBLP:conf/emnlp/Kim14} is CNN-based recommendation model. (2) \textit{Trans} \cite{DBLP:conf/nips/VaswaniSPUJGKP17} is a Transformer-based response generation model. (3) \textit{KBRD} \cite{DBLP:conf/emnlp/ChenLZDCYT19} is a knowledge-based CRS that enhances user preference with a KG. (4) \textit{KGSF} \cite{DBLP:conf/kdd/ZhouZBZWY20} is a KG-based CRS aligning the semantic space of two KGs. (5) \textit{RevCore} \cite{DBLP:conf/acl/LuBSMCWH21} is a review-enhanced CRS. (6) \textit{CRFR} \cite{DBLP:conf/emnlp/ZhouWHH21} is a fragments reasoning-based CRS. (7) \textit{C$^2$-CRS} \cite{DBLP:journals/corr/abs-2201-02732} is a contrastive learning-based CRS. (8) We design \textit{ACRG} as a variant of DICR and removes all imitation components and rich interaction between two modules from DICR.

\paragraph{Implementation Details}
\label{sec:implementation}
We implemented our model with Pytorch. In the recommendation module, the history length $H=1$ and the maximum length of the reasoning path is 3. The maximum action space is 250. We trained the KG with the embedding size 128. The rewards weights are $\alpha=\gamma=0.006$, $\beta=0.001$. The conversation module receives $N_{p}=10$ recommendation paths. All encoders and decoders have 2-layers with 800 hidden units for each layer. The word embedding is initialized with word2vec and size 300. We used the Adam optimizer \cite{DBLP:journals/corr/KingmaB14}, the batch size is 32 and the learning rate is 0.0001. We trained our model with four steps. We first trained the model to minimize the $L_{R E C}$ loss, then minimized the BOW loss and BCE loss for pre-training knowledge imitation and semantic imitation components, and then minimized the $L_{G E N}$ loss. Finally, we jointly trained the whole model. 

\subsection{Evaluation on Recommendation}

\begin{table}[t]
\centering
\resizebox{.8\columnwidth}{!}{
\begin{tabular}{l c c c}
\hline
    Models & Recall@1 & Recall@10 & Recall@25 \\
\hline
    TextCNN & 0.059 & 0.177 & 0.235 \\
    KBRD & 0.104 & 0.407 & 0.490 \\
    KGSF & 0.119 & 0.436 & 0.523 \\
    RevCore & 0.124 & 0.432 & 0.516 \\
    CRFR & 0.130 & 0.458 & 0.543 \\
    C$^2$-CRS & 0.112 & 0.465 & 0.541 \\
    ACRG & 0.185 & 0.490 & 0.629 \\
\hline
    \textbf{DICR} & $\textbf{0.211}^\ast$ & $\textbf{0.511}^\ast$ & $\textbf{0.643}^\ast$ \\
\hline
    w/o PI & 0.203 & 0.497 & 0.635 \\
    w/o KI & 0.201 & 0.494 & 0.632 \\
    w/o SI & 0.200 & 0.500 & 0.631 \\
\hline
\end{tabular}}
\caption{Overall evaluation on recommendation. w/o refers to removing the component from DICR. “$\ast$” indicates the statistical significance for $p \textless 0.001$ compared with the best baseline (t-test with p-value $\textless$ 0.001).}
\label{recommendation}
\end{table}

In recommendation evaluation, we use Recall@K (K=1,10,25) indicating whether the top-k predicted items include the golden recommendation item.

\paragraph{Overall Evaluation}

As in Table \ref{recommendation}, DICR outperforms all the baselines significantly, benefiting from using rich dual imitation signals as the rewards for the recommendation agent. Compared with the best results of CRFR and C$^2$-CRS, DICR achieves 62.3\%, 9.9\%, and 18.4\% improvements on three metrics. Despite substantial progress achieved by extra knowledge (i.e., KBRD, KGSF, RevCore, C$^2$-CRS) and fragments reasoning (i.e., CRFR), their performance is still inferior to ACRG indicating that black-box preference representation is a sub-optimal interest expression scheme.

\paragraph{Ablation Study of Dual Imitation}

We separately remove path imitation, knowledge imitation and semantic imitation from DICR to examine their contribution, called w/o PI, w/o KI and w/o SI.
In Table \ref{recommendation}, all imitation components contribute to the recommendation performance, with the rewards within module (PI) or across modules (KI and SI) for the reasoning policy learning. 
On the one hand, the conversational rewards (i.e., from KI and SI) are used as the alignment signals and guide the recommender to learn user interest shift policies. On the other hand, the dual-reward-reinforced (i.e., from PI, KI and SI) recommendation paths serve as the alignment signals in turn improve the conversation by promoting the positive cyclic learning of bidirectional interaction with dual imitation.

\subsection{Evaluation on Conversation}

\paragraph{Overall Evaluation}

\begin{table}[t]
\centering
\resizebox{.8\columnwidth}{!}{
\begin{tabular}{l c c c c c}
\hline
    Models & Bleu-1 & Bleu-2 & Dist-1 & Dist-2 & F1 \\
\hline
    Trans & 0.388 & 0.309 & 0.027 & 0.103 & 0.050 \\
    KBRD & 0.408 & 0.324 & 0.055 & 0.162 & 0.108 \\
    KGSF & 0.416 & 0.330 & 0.062 & 0.203 & 0.123 \\
    RevCore & 0.409 & 0.323 & 0.057 & 0.195 & 0.112 \\
    CRFR & 0.421 & 0.334 & 0.064 & 0.208 & 0.135 \\
    C$^2$-CRS & 0.417 & 0.331 & \textbf{0.065} & \textbf{0.209} & 0.145 \\
    ACRG & 0.422 & 0.326 & 0.054 & 0.161 & 0.270 \\
\hline
    \textbf{DICR} & $\textbf{0.478}^\ast$ & $\textbf{0.366}^\ast$ & 0.059 & 0.183 & $\textbf{0.319}^\ast$ \\
\hline
    w/o PI & 0.467 & 0.357 & 0.057 & 0.177 & 0.308 \\
    w/o KI & 0.464 & 0.357 & 0.055 & 0.171 & 0.293 \\
    w/o SI & 0.466 & 0.358 & 0.057 & 0.175 & 0.302 \\
    w/o FG & 0.458 & 0.352 & 0.056 & 0.168 & 0.305 \\
    w/o BI & 0.439 & 0.338 & 0.058 & 0.174 & 0.288 \\
\hline
\end{tabular}}
\caption{Overall evaluation on conversation. w/o and “$\ast$” have the same meaning as those in Table 1.}
\label{conversationoverall}
\end{table}

To evaluate the overall performance of the conversation, we use BLEU-1/2 and Distinct-1/2 (Dist-1/2) to evaluate the quality and diversity of the generated responses. F1 is the F1-score measuring how well the responses contain golden knowledge. 
In Table \ref{conversationoverall}, DICR outperforms all baselines significantly in most metrics.
DICR achieves 13.3\%, 9.6\% improvements on Bleu-1/2 compared to the best baselines, which supports the effectiveness of our method, i.e., aligning the recommendation reasoning and conversation process. DICR achieves the best result on F1, demonstrating that explicit recommendation paths (i.e., DICR, ACRG) are superior to implicit embedding semantics (i.e., baselines except for ACRG) in guiding the generation of knowledge-rich responses.

\paragraph{Hit and Explainability}

Accurate recommendations with coherent explanations is one of our main contributions. We propose \textbf{“Hit”} to measure the recommendation success rate in conversation. Hit is the hit rate at which recommended items in the golden response are included in the generated response. The explainability of the response is evaluated by logically linked entity pairs which are necessary for coherent explanation. Specifically, \textbf{“Inter”} counts the entity links between context and response, which evaluates contextually coherent explanation across context and response. \textbf{“Inner”} counts the entity links within the response, which evaluates self-consistent explanation in response. \textbf{“G”} counts the entity links that can be matched in KG and thus evaluates the explanations with global KG knowledge. \textbf{“P”} counts the entity links that can be matched in the recommendation paths in KG and thus evaluates how well the recommendation paths support the explanation generation. Finally, we have four combined indicators, \textbf{“G-Inter, G-Inner, P-Inter, P-Inner”}, e.g., \textbf{
“G-Inter”} evaluates the coherent explanation according to KG knowledge.

\begin{table}[t]
\centering
\resizebox{.8\columnwidth}{!}{
\begin{tabular}{l c c c c c}
\hline
    Models & Hit & G-Inter & G-Inner & P-Inter & P-Inner \\
\hline
    KBRD & 0.250 & 0.639 & 0.194 & - & - \\
    KGSF & 0.261 & 0.662 & 0.201 & - & - \\
    RevCore & 0.245 & 0.670 & 0.222 & - & - \\
    CRFR & 0.293 & 0.691 & 0.247 & 0.485 & 0.134 \\
    C$^2$-CRS & 0.310 & 0.695 & 0.319 & - & - \\
    ACRG & 0.318 & 0.548 & 0.288 & 0.527 & 0.279 \\
\hline
    \textbf{DICR} & $\textbf{0.426}^\ast$ & $\textbf{0.720}^\ast$ & $\textbf{0.348}^\ast$ & $\textbf{0.689}^\ast$ & $\textbf{0.330}^\ast$ \\
\hline
    w/o PI & 0.405 & 0.685 & 0.320 & 0.657 & 0.309 \\
    w/o KI & 0.386 & 0.680 & 0.342 & 0.651 & 0.327 \\
    w/o SI & 0.403 & 0.687 & 0.327 & 0.663 & 0.319 \\
    w/o FG & 0.385 & 0.667 & 0.263 & 0.641 & 0.256 \\
    w/o BI & 0.354 & 0.592 & 0.298 & 0.570 & 0.287 \\
\hline
\end{tabular}}
\caption{Evaluation on hit and explainability. w/o and “$\ast$” have the same meaning as those in Table 1.}
\label{conversationhit}
\end{table}

In Table \ref{conversationhit}, DICR outperforms all baselines and obtains a significant improvement on Hit and explainability.
First, DICR achieves 34\% improvements on Hit compared to the best ACRG, which verifies the effectiveness of the dual imitation mechanism for aligning the consistent behavior of the recommendation reasoning and conversation process. 
Second, DICR obtains 3.6\% and 9.1\% gains on G-Inter and G-Inner compared to the best C$^2$-CRS, which shows that DICR prefers to generate logically coherent explanations within responses and across context and response.
Third, DICR improves 30.7\% and 18.3\% on P-Inter and P-Inner compared to the best ACRG, which indicates that the conversation side of the dual imitation (i.e., KI and SI) can effectively identify and integrate recommendation paths for response generation.

\paragraph{Human Evaluation}

In human evaluation, we randomly sampled 200 contexts. Each context is associated with eight responses from eight comparison models, respectively. Six well-educated annotators evaluate each response with four scores: fluency, coherence, informativeness, and explainability. Fluency and coherence evaluate the language quality of responses. Informativeness evaluates whether the response incorporates rich knowledge. Explainability evaluates whether the response explains the reason for the recommended item. The scores are set to $\{0,1,2\}$. The model name is masked during the evaluation for a fair comparison. The Fleiss’ kappa \cite{fleiss1971measuring} measures the agreement among the annotators. As the results in Table \ref{human}, the superior of DICR on all indicators support the observations from automatic evaluations.

\begin{table}[t]
\centering
\resizebox{.75\columnwidth}{!}{
\begin{tabular}{l c c c c}
\hline
    Models & Flu. & Coher. & Inform. & Explain. \\
\hline
    Trans & 1.644 & 1.302 & 1.002 & 0.750 \\
    KBRD & 1.688 & 1.328 & 1.235 & 0.875 \\
    KGSF & 1.665 & 1.351 & 1.241 & 0.895 \\
    RevCore & 1.692 & 1.383 & 1.248 & 0.912 \\
    CRFR & 1.696 & 1.386 & 1.262 & 0.940 \\
    C$^2$-CRS & 1.673 & 1.342 & 1.234 & 0.900 \\
    ACRG & 1.790 & 1.608 & 1.045 & 0.950 \\
\hline
    \textbf{DICR} & $\textbf{1.815}^\ast$ & $\textbf{1.633}^\ast$ & $\textbf{1.300}^\ast$ & $\textbf{1.058}^\ast$ \\
\hline
    kappa & 0.566 & 0.518 & 0.521 & 0.490 \\
\hline
\end{tabular}}
\caption{Human evaluation on conversation. “Flu.”, “Coher.”, “Inform.”, “Explain.” denote fluency, coherence, informativeness and explainability. The agreement ratio $kappa \in [0.41,0.6]$ denotes the moderate agreement. “$\ast$” indicates t-test with p-value $\textless$ 0.05.}
\label{human}
\end{table}

\begin{figure*}[t]
\centering
\includegraphics[width=0.94\textwidth]{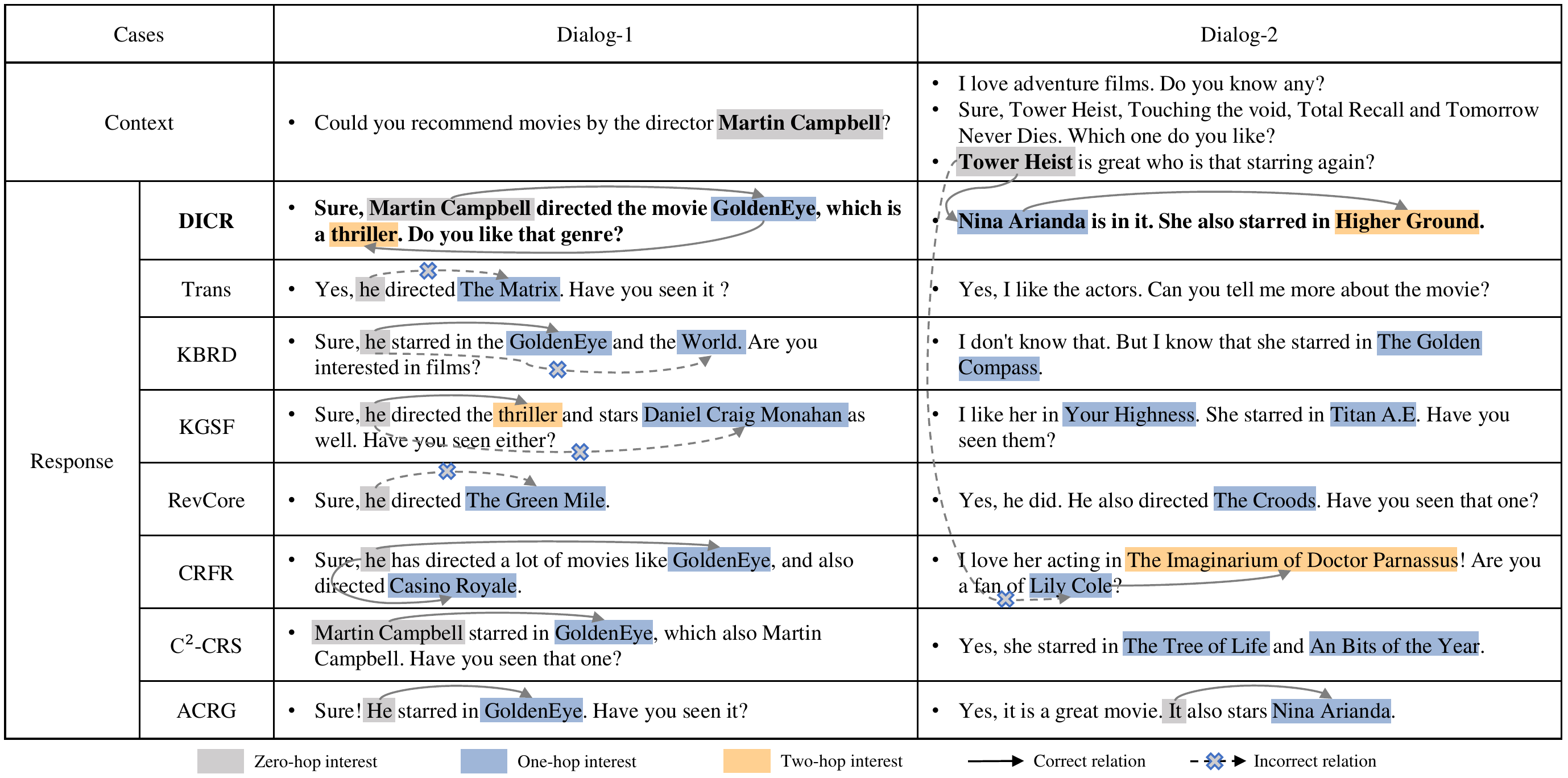} % Reduce the figure size so that it is slightly narrower than the column.
\caption{Cases generated by different models, indicting multi-hop entities and correct/incorrect relations. }
\label{casestudy}
\end{figure*}

\paragraph{Ablation Study}

In Tables \ref{conversationoverall} and \ref{conversationhit}: 
(1) We separately remove the path imitation, knowledge imitation, and semantic imitation to examine their contribution, namely w/o PI, w/o KI, and w/o SI, respectively. In the results, the path imitation mainly benefits the inner coherent of explainability (“G/P-Inner”), which verifies its designed advantage to indirectly guide the explanation logic by accurate recommendation paths. The knowledge imitation mainly benefits recommendation hit (“Hit”), inter coherent of recommendation (“G/P-Inter”) and distinct of responses (“Dist-1/2”), which verifies its designed advantage to refine the distribution of recommendation paths for accurate recommendation and to encourage diverse explanations in response. The semantic imitation also mainly benefits inner coherent of explainability and is more important to inter coherent of recommendation than path imitation, which verifies its designed advantages to improve the semantic of response by promoting inner and inter coherent. 
(2) We remove the fusion gate, namely w/o FG. The results show that the dynamic information fusion mechanism achieves an impressive enhancement.
(3) We remove the bidirectional improvement in the training, namely w/o BI. The results indicate that the tightly information interaction between recommendation and conversation with the alignment signals of dual imitation as the bridge is crucial to the overall performance.

\subsection{Case Study}
\label{sec:casestudy}

In Figure \ref{casestudy}, two cases from eight models are selected, among which DICR has two advantages:
(1) The items recommended by DICR is more accurate and likely to have explicit multi-hop relation with the items mentioned by the user, being consistent to “Hit” and “G/P-Inter” in Table \ref{conversationhit}, e.g., in Dialog-2, “Higher Ground” in response and “Tower Heist” in context share the actor “Nina Arianda.” This is evidence of improving recommendation by tracking user interest shift in conversation, which mainly benefits from the path imitation and the knowledge imitation, as verified by ablation study in Table \ref{conversationhit}; 
(2) DICR naturally tells the items' relation as an explanation, being consistent to “G/P-Inner” in Table \ref{conversationhit} and “Explain.” in Table \ref{human}, e.g., in Dialog-1, the director “Martin Campbell”, the movie “GoldenEye” and the genre “thriller” derive from a recommendation path with coherent relation. This is evidence of improving conversation by involving recommendation path as an explanation, which mainly benefits from the semantic imitation, as verified in Table \ref{conversationhit}. 

\section{Conclusions}

We propose DICR, which adopts the dual imitation to align CRS's recommendation and conversation behavior explicitly. Using recommendation paths and conversational rewards as alignment signals for tight interaction between recommendation and conversation, DICR achieves accurate recommendations and coherent explanations in generated responses. The effectiveness of DICR is verified by designed novel explainability evaluations together with human and existing automatic metrics. 

\section*{Limitations}
We discuss two main limitations of this work which can be further studied in future work. The first one is the reliance of explicit knowledge in knowledge graph. Although using knowledge graph is a common advantage of most current CRS studies, and explicit relations between entities leads to effective and reliable reasoning for recommendation, there are still a large amount of implicit knowledge in unstructured resources which cannot be extracted as explicit triplet, e.g., the multidimensional similarity between entities, but can be further extra supplement to dialog context. 

The second one is the task of next-turn recommendation. As the main contribution of this work, although the modeling of user interest shift significantly improve the performance of making recommendation in next-turn response, the user interest shift modeling can also naturally help us to guide the user interests towards proper recommendation through smooth and persuasive multi-turn conversation with users. To address this limitation, in the future, we will extend the idea to align the KG-based reasoning and conversation process towards long-term global goal instead of local target. 

\section*{Ethics Consideration}
All models in this paper are trained on public corpus. The OpenDialKG \cite{DBLP:conf/acl/MoonSKS19} dataset do not contain personal information and unethical language. We also ensure the anonymization of the human evaluation. We belieive that this work honors the ethical code of EMNLP.

\section*{Acknowledgements}

This work was supported by National Natural Science Foundation of China (62272340, 61876128, 61876129, 62276187, 61976154, 61402323), State Key Laboratory of Communication Content Cognition (Grant No.A32003).

\bibliography{anthology,crs}
\bibliographystyle{acl_natbib}

\clearpage

\appendix

\section{Dataset}
\label{sec:dataset}

The statistics of OpenDialKG after preprocessing are in Table \ref{dataset}.

We did not employ other CRS datasets. The reason is that compared with OpenDialKG, in other datasets like REDIAL \cite{DBLP:conf/nips/LiKSMCP18}, dialogs mention the recommended items without rich related information and tend to mention only movie names rather than an in-depth discussion on the movie preference, which is considered as the recommended explanation in this paper. As reported in the \textit{CRFR} \cite{DBLP:conf/emnlp/ZhouWHH21}, OpenDialKG's advantages improve the performance of \textit{CRFR} and the compared CRS baselines in our experiments.

\begin{table}[t]
\centering
\resizebox{.8\columnwidth}{!}{
\begin{tabular}{c l c}
\hline
    \multirow{5}{*}{\makecell[c]{Corpus \\ Info.}} 
    & \#Domain & Movie,Book \\
    & \#Dialogues & 15,673 \\
    & \#Turns & 91,209 \\
    & \#Split Ratio & 7:1.5:1.5 \\
\hline
    \multirow{3}{*}{\makecell[c]{KG Info.}} 
    & \#Entities & 100,813 \\
    & \#Relations & 1,358 \\
    & \#Triplets & 1,190,658 \\
\hline
\end{tabular}}
\caption{Statistics of our datasets after preprocessing.}
\label{dataset}
\end{table}

\begin{figure*}
  \setlength{\abovecaptionskip}{5pt}
  \centering
    \begin{subfigure}{.32\textwidth}
        \centering
        \setlength{\abovecaptionskip}{5pt}
        \includegraphics[width=1.\textwidth]{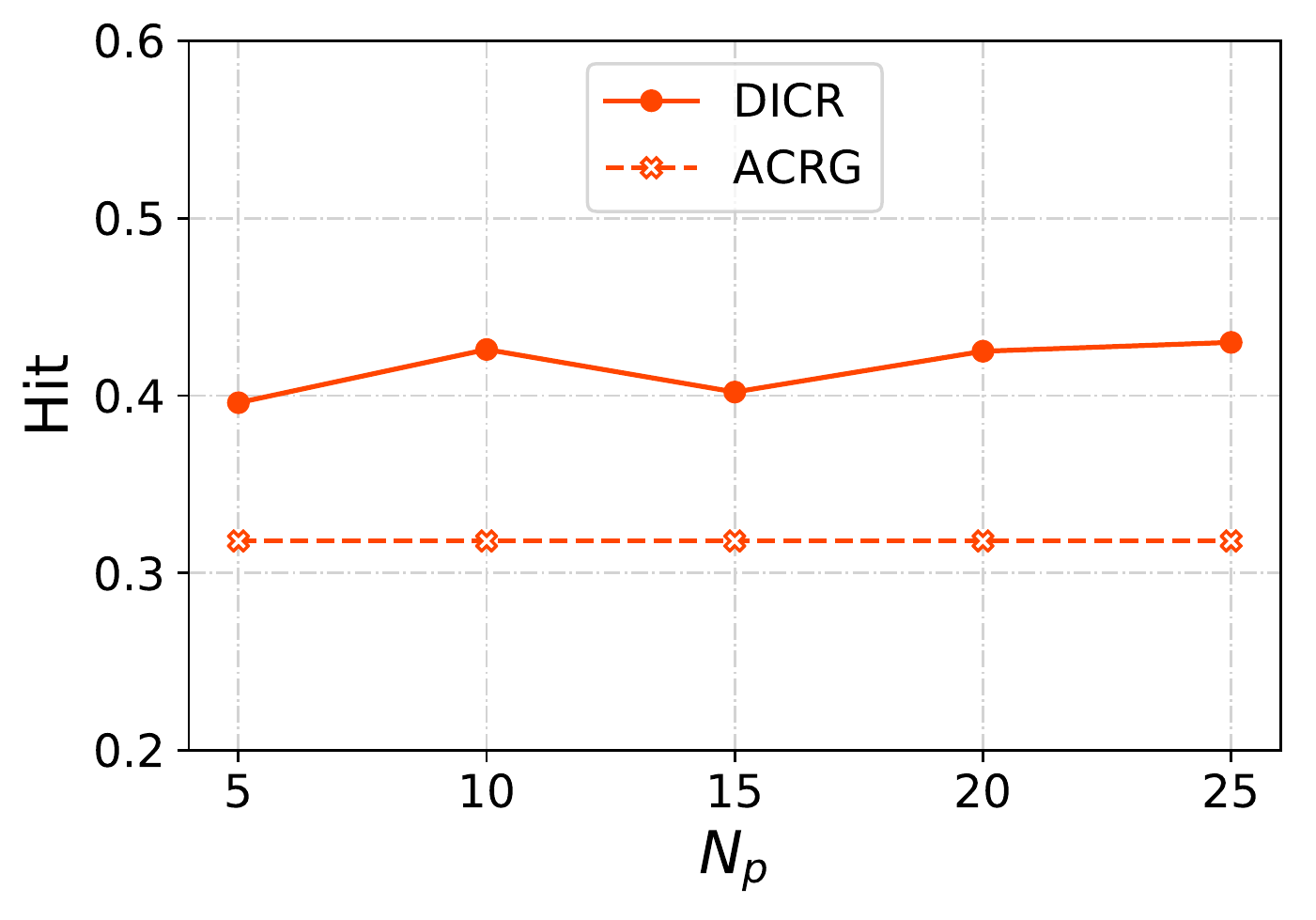}
        \caption{Hit}
        \label{hit}
    \end{subfigure}
    \begin{subfigure}{.32\textwidth}
        \centering
        \setlength{\abovecaptionskip}{5pt}
        \includegraphics[width=1.\textwidth]{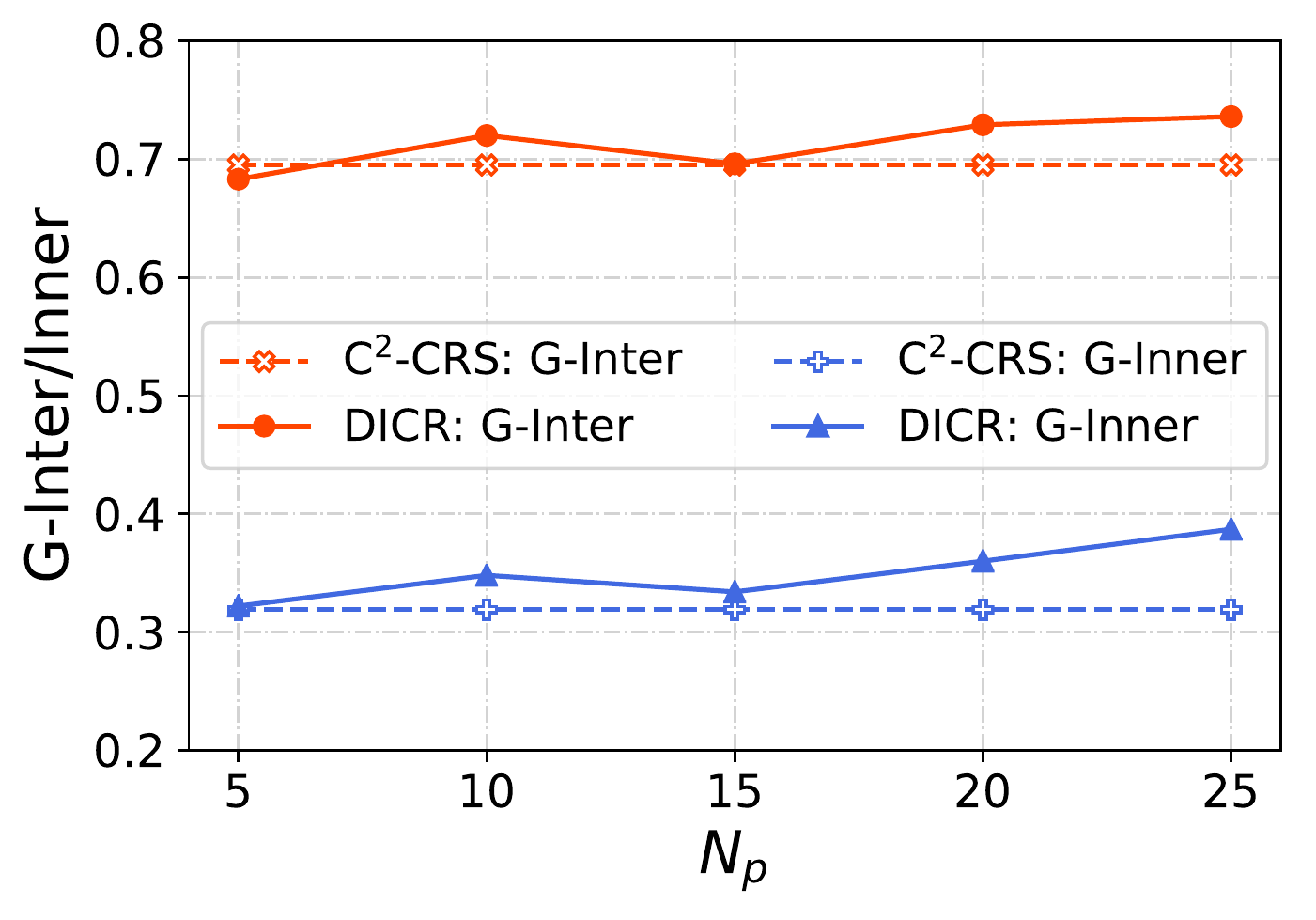}
        \caption{G-Inter/Inner}
        \label{ginter}
    \end{subfigure}
    \begin{subfigure}{.32\textwidth}
        \centering
        \setlength{\abovecaptionskip}{5pt}
        \includegraphics[width=1.\textwidth]{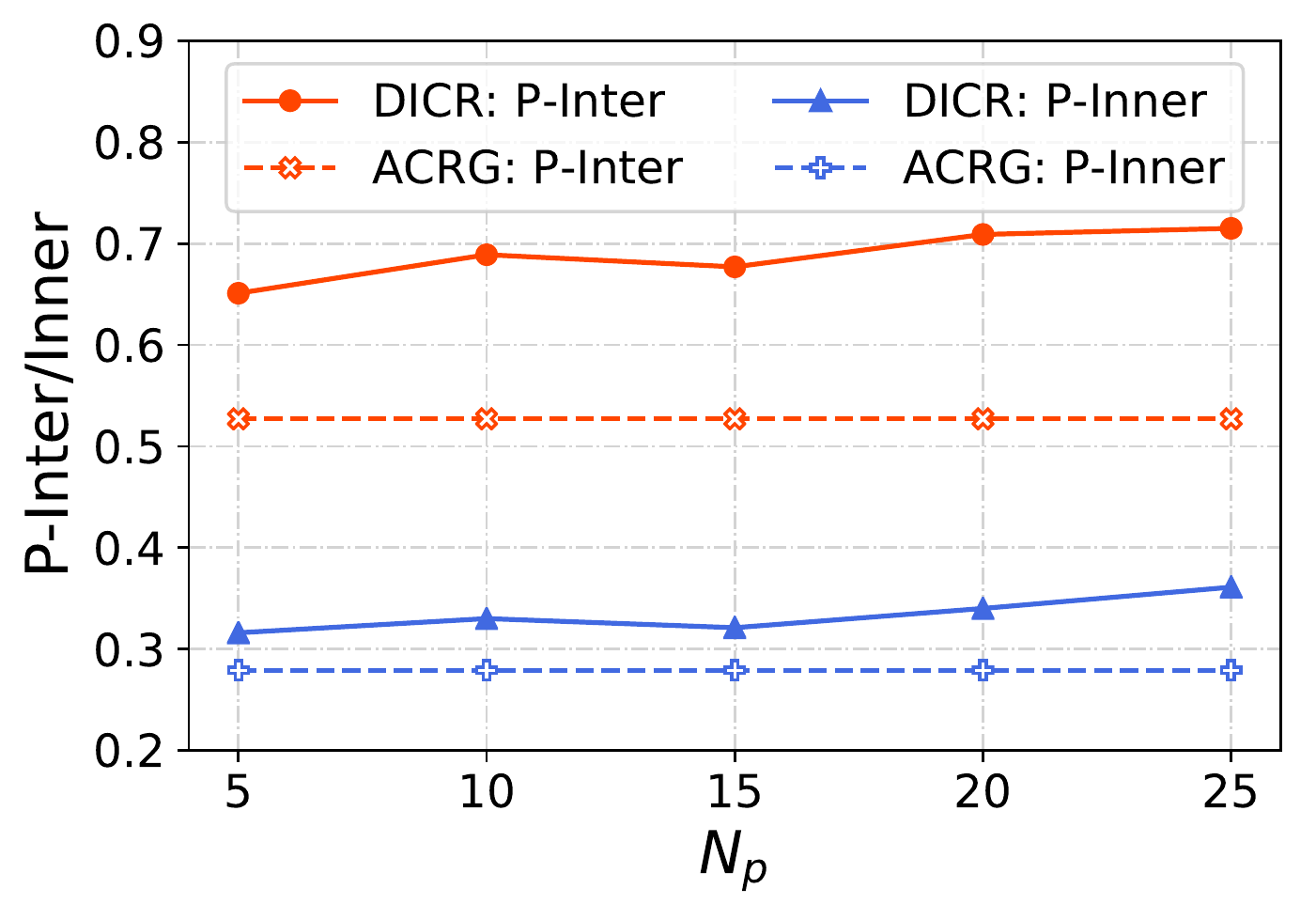}
        \caption{P-Inter/Inner}
        \label{pinter}
    \end{subfigure}
  \caption{The influence of the number of recommendation paths on Hit and Explainability. As the number of recommendation paths increases, DICR improves on Hit and Explainability metrics and outperforms the best baseline on most cases.}
  \label{pathnumber}
\end{figure*}

\section{Analysis of the Number of Recommendation Paths}
\label{sec:pathnumber}
We analyze the influence of the number of recommendation paths on Hit and explainability, and Figure \ref{pathnumber} presents the results. 

First, as shown in Figure \ref{hit}, with the increase of the number of recommendation paths which are used as the alignment signals for the recommendation side of the dual imitation, the Hit scores slightly improve in fluctuations. This indicates that the conversation side of the dual imitation can effectively identify the golden recommendation paths and prompt the conversation process to align the recommendation reasoning. 

Second, in Figure \ref{ginter} and \ref{pinter}, G-Inter/Inner and P-Inter/Inner both improve distinctly as the number of paths increases. This improvement is attributed to the advantage that knowledge imitation and semantic imitation endow the DICR with discerning and integrating the coherent knowledge in the recommendation paths as the recommended explanation in response. This advantage aligns the recommendation reasoning to explanation generation, which helps the model refine the discerned knowledge and display them in the generated response.

\end{document}